\documentclass{article}
\pdfoutput=1

 \PassOptionsToPackage{numbers, compress}{natbib}

\usepackage[preprint]{nips_2018}

\usepackage[utf8]{inputenc} 
\usepackage[T1]{fontenc}    
\usepackage{hyperref}       
\usepackage{url}            
\usepackage{booktabs}       
\usepackage{amsfonts}       
\usepackage{nicefrac}       
\usepackage{microtype}      

\usepackage{graphicx}
\usepackage{algorithm}
\usepackage[noend]{algpseudocode}
\usepackage{amsmath}
\usepackage{multirow}
\usepackage{tabularx}
\usepackage{wrapfig}
\usepackage{lipsum}
\usepackage{array}
\usepackage{subcaption}
\usepackage{mathtools}
\usepackage{icomma}
\usepackage{siunitx}
\usepackage{csquotes}
\usepackage{soul}
\usepackage{cleveref}
\usepackage{tikz}
\usetikzlibrary{bayesnet}

\usepackage{amsmath,amsthm,amssymb}

\newcommand\cut[1]{}

\newcolumntype{C}[1]{>{\centering\arraybackslash}m{#1}}
\newcolumntype{R}[1]{>{\raggedleft\arraybackslash}m{#1}}

\newcommand{\KLpq}[2]{\mathbb{KL}\left({#1}||{#2}\right)}

\newcommand{\myvec}[1]{\mathbf{#1}}

\newcommand{\va}{\myvec{a}}

\newcommand{\vs}{\myvec{s}}

\newcommand{\vu}{\myvec{u}}

\newcommand{\vx}{\myvec{x}}

\newcommand{\vz}{\myvec{z}}

\newcommand{\E}{\mathbb{E}}

\newcommand{\diag}{\mbox{diag}}

\newcommand{\be}{\begin{equation}}
\newcommand{\ee}{\end{equation}}
\newcommand{\bea}{\begin{eqnarray}}
\newcommand{\eea}{\end{eqnarray}}
\newcommand{\beaa}{\begin{eqnarray*}}
\newcommand{\eeaa}{\end{eqnarray*}}

\DeclareMathAlphabet{\mathpzc}{OT1}{pzc}{m}{n}

\newcommand{\utext}[2]{\underbrace{#1}_{\text{#2}}}

\graphicspath{{figures/}}
\DeclarePairedDelimiterX{\divx}[2]{\big(}{\big)}{%
  #1\;\delimsize\|\;#2%
}

\usepackage{xspace}
\usepackage[subtle,tracking=normal]{savetrees}

\renewcommand{\L}{\mathcal{L}}
\newcommand{\enc}{\text{enc}}

\newcommand{\betavae}{\ensuremath{\beta}-VAE\xspace}

\newcommand{\ourawesomemodel}{VASE\xspace}
\newcommand{\flying}{moving\xspace}
\newcommand{\fmnist}{Fashion\xspace}

\newboolean{draftmode}
\setboolean{draftmode}{false}

\usepackage{color}
\definecolor{celestialblue}{rgb}{0.29, 0.59, 0.82}
\newcommand{\mycomment}[3]{{\textcolor{#3}{[#1 #2]}}}
\newcommand{\ihmarker}{{\textcolor{blue}{\ensuremath{^{\textsc{I}}_{\textsc{H}}}}}}
\newcommand{\aamarker}{{\textcolor{red}{\ensuremath{^{\textsc{A}}_{\textsc{A}}}}}}
\newcommand{\cbmarker}{{\textcolor{orange}{\ensuremath{^{\textsc{C}}_{\textsc{B}}}}}}
\newcommand{\lmmarker}{{\textcolor{green}{\ensuremath{^{\textsc{L}}_{\textsc{M}}}}}}
\newcommand{\nwmarker}{{\textcolor{magenta}{\ensuremath{^{\textsc{N}}_{\textsc{W}}}}}}
\newcommand{\temarker}{{\textcolor{teal}{\ensuremath{^{\textsc{T}}_{\textsc{E}}}}}}
\newcommand{\almarker}{{\textcolor{pink}{\ensuremath{^{\textsc{A}}_{\textsc{L}}}}}}

\ifthenelse{\boolean{draftmode}}{
 \newcommand{\ih}[1]{\mycomment{\ihmarker}{#1}{blue}}
 \newcommand{\aach}[1]{\mycomment{\aamarker}{#1}{red}}
 \newcommand{\cb}[1]{\mycomment{\cbmarker}{#1}{orange}}
 \newcommand{\lm}[1]{\mycomment{\lmmarker}{#1}{green}}
 \newcommand{\nw}[1]{\mycomment{\nwmarker}{#1}{magenta}}
 \newcommand{\te}[1]{\mycomment{\temarker}{#1}{teal}}
 \newcommand{\al}[1]{\mycomment{\almarker}{#1}{pink}}

}{
 \newcommand{\ih}[1]{}
 \newcommand{\aach}[1]{}
 \newcommand{\cb}[1]{}
 \newcommand{\lm}[1]{}
 \newcommand{\nw}[1]{}
 \newcommand{\te}[1]{}
 \newcommand{\al}[1]{}
}

\makeatletter
\def\BState{\State\hskip-\ALG@thistlm}
\makeatother

\title{
Life-Long Disentangled Representation Learning with Cross-Domain Latent Homologies
}

\author{
 Alessandro~Achille, Tom~Eccles, Loic~Matthey, Christopher~P~Burgess, \\ 
 \textbf{Nick~Watters, Alexander~Lerchner, Irina~Higgins } \\
 UCLA, DeepMind\\
 achille@cs.ucla.edu,\\
 \texttt{\{eccles,lmatthey,cpburgess,nwatters,lerchner,irinah\}@google.com}
}

\begin{document}
\maketitle

\begin{abstract}
Intelligent behaviour in the real-world requires the ability to acquire new knowledge from an ongoing sequence of experiences while preserving and reusing past knowledge. We propose a novel algorithm for unsupervised representation learning from piece-wise stationary visual data: Variational Autoencoder with Shared Embeddings (\ourawesomemodel). Based on the Minimum Description Length principle, \ourawesomemodel automatically detects shifts in the data distribution and allocates spare representational capacity to new knowledge, while simultaneously protecting previously learnt representations from catastrophic forgetting. Our approach encourages the learnt representations to be disentangled, which imparts a number of desirable properties: \ourawesomemodel can deal sensibly with ambiguous inputs, it can enhance its own representations through imagination-based exploration, and most importantly, it exhibits semantically meaningful sharing of latents between different datasets. Compared to baselines with entangled representations, our approach is able to reason beyond surface-level statistics and perform semantically meaningful cross-domain inference. 
\end{abstract}


\section{Introduction}
\label{sec_intro}
A critical feature of biological intelligence is its capacity for \emph{life-long learning} \cite{Cichon_Gan_2015} -- the ability to acquire new knowledge from a sequence of experiences to solve progressively more tasks, while maintaining performance on previous ones. This, however, remains a serious challenge for current deep learning approaches. While current methods are able to outperform humans on many individual problems \cite{Silver_etal_2016, Mnih_etal_2015, He_etal_2015}, these algorithms suffer from \emph{catastrophic forgetting} \cite{French_1999, McClelland_etal_1995, McCloskey_Cohen_1989, Ratcliff_1990, Goodfellow_etal_2013}. Training on a new task or environment can be enough to degrade their performance from super-human to chance level \cite{Rusu_etal_2016}. Another critical aspect of life-long learning is the ability to sensibly reuse previously learnt representations in new domains (\emph{positive transfer}). For example, knowing that strawberries and bananas are not edible when they are green could be useful when deciding whether to eat a green peach in the future. Finding semantic homologies between visually distinctive domains can remove the need to learn from scratch on every new environment and hence help with data efficiency -- another major drawback of current deep learning approaches \cite{Garnelo_etal_2016, Lake_etal_2016}. 

But how can an algorithm maximise the informativeness of the representation it learns on one domain for positive transfer on other domains without knowing a priori what experiences are to come? One approach might be to capture the important structure of the current environment in a maximally compact way (to preserve capacity for future learning). Such learning is likely to result in positive transfer if future training domains share some structural similarity with the old ones. This is a reasonable expectation to have for most natural (non-adversarial) tasks and environments, since they tend to adhere to the structure of the real world (e.g. relate to objects and their properties) governed by the consistent rules of chemistry or physics. A similar motivation underlies the Minimum Description Length (MDL) principle \cite{rissanen1978modeling} and disentangled representation learning \cite{Bengio_etal_2013}. 

\begin{figure}
    \centering
    \includegraphics[width=0.9\textwidth]{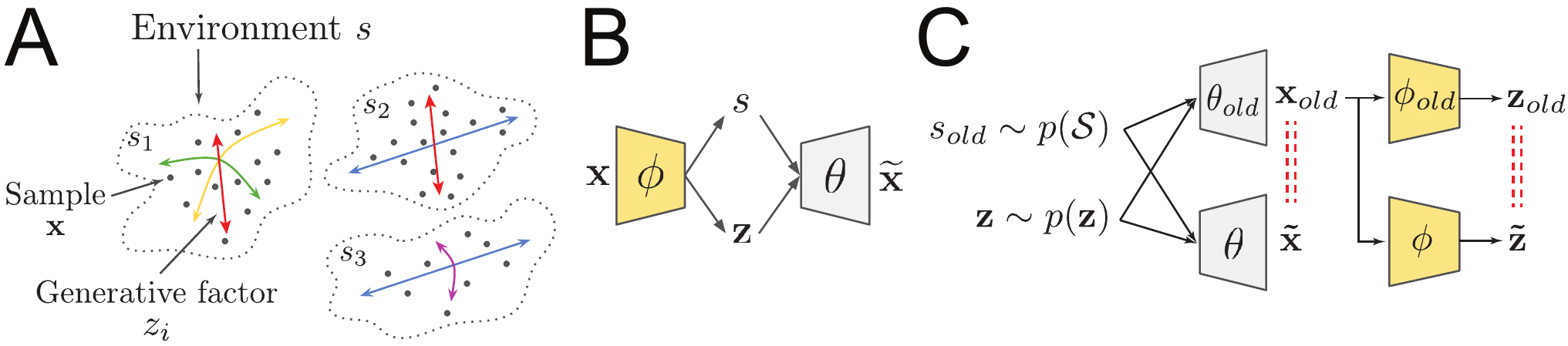}
    \caption{
    \textbf{A}: Schematic representation of the life-long learning data distribution. Each dataset/environment corresponds to a cluster $s$. Data samples $\vx$ constituting each cluster can be described by a local set of coordinates (data generative factors $z_n$). Different clusters may share some data generative factors. \textbf{B}: \ourawesomemodel model architecture \textbf{C}: ConContinSchematic of the ``dreaming'' feedback loop. We use a snapshot of the model with the old parameters ($\phi_{\text{old}}$, \ $\theta_{\text{old}})$ to generate an imaginary batch of data $\vx_{\text{old}}$ for a previously experienced dataset $s_{\text{old}}$. While learning in the current environment, we ensure that the representation is still consistent on the hallucinated ``dream'' data, and can reconstruct it (see red dashed lines).
    }
    \label{fig:network}
\end{figure}

Recent state of the art approaches to unsupervised disentangled representation learning \cite{Higgins_etal_2017, Burgess_etal_2017, Kim_Mnih_2017, Kumar_etal_2018} use a modified Variational AutoEncoder (VAE) \cite{Kingma_Welling_2014, Rezende_etal_2014} framework to learn a representation of the data generative factors. These approaches, however, only work on independent and identically distributed (IID) data from a single visual domain. This paper extends this line of work to life-long learning from piece-wise stationary data, exploiting this setting to learn shared representations across domains where applicable. The proposed Variational Autoencoder with Shared Embeddings (\ourawesomemodel, see \cref{fig:network}B) automatically detects shifts in the training data distribution and uses this information to allocate spare latent capacity to novel dataset-specific disentangled representations, while reusing previously acquired representations of latent dimensions where applicable. We use latent masking and a generative ``dreaming'' feedback loop (similar to \cite{ramapuram2017lifelong, Shin_etal_2017, Seff_etal_2017, Ans_Rousset_1997}) to avoid catastrophic forgetting. Our approach outperforms \cite{ramapuram2017lifelong}, the only other VAE based approach to life-long learning we are aware of. Furthermore, we demonstrate that the pressure to disentangle endows \ourawesomemodel with a number of useful properties: 1) dealing sensibly with ambiguous inputs; 2) learning richer representations through imagination-based exploration; 3) performing semantically meaningful cross-domain inference by ignoring irrelevant aspects of surface-level appearance.

\section{Related work}
\label{sec:related_work}
The existing approaches to continual learning can be broadly separated into three categories: data-, architecture- or weights-based. The data-based approaches augment the training data on a new task with the data collected from the previous tasks, allowing for simultaneous multi-task learning on IID data \cite{Espeholt_etal_2018, Robins_1995, Ratcliff_1990, McClelland_etal_1995, furlanello2016active}. The architecture-based approaches dynamically augment the network with new task-specific modules, which often share intermediate representations to encourage positive transfer \cite{Rusu_etal_2016, Parisotto_etal_2015,Ruvolo_Eaton_2013}. Both of these types of approaches, however, are inefficient in terms of the memory requirements once the number of tasks becomes large. The weights-based approaches do not require data or model augmentation. Instead, they prevent catastrophic forgetting by slowing down learning in the weights that are deemed to be important for the previously learnt tasks \cite{Kirkpatrick_etal_2017, Zenke_etal_2017, Nguyen_etal_2018}. This is a promising direction, however, its application is limited by the fact that it typically uses knowledge of the task presentation schedule to update the loss function after each switch in the data distribution. 

Most of the continual learning literature, including all of the approaches discussed above, have been developed in task-based settings, where representations are learnt implicitly. While deep networks learn well in such settings \cite{achille2017emergence, shwartz2017opening}, this often comes at a cost of reduced positive transfer. This is because the implicitly learnt representations often overfit to the training task by discarding information that is irrelevant to the current task but may be required for solving future tasks \cite{achille2017emergence, achille2018information, achille2017separation, shwartz2017opening, Higgins_etal_2017b}. The acquisition of useful representations of complex high-dimensional data without task-based overfitting is a core goal of unsupervised learning. Past work \cite{achille2018information, alemi2016deep, Higgins_etal_2017} has demonstrated the usefulness of information-theoretic methods in such settings. These approaches can broadly be seen as efficient implementations of the Minimum Description Length (MDL) principle for unsupervised learning \cite{rissanen1978modeling, grunwald2007minimum}. The representations learnt through such methods have been shown to help in transfer scenarios and with data efficiency for policy learning in the Reinforcement Learning (RL) context \cite{Higgins_etal_2017b}. These approaches, however, do not immediately generalise to non-stationary data. Indeed, life-long unsupervised representation learning is relatively under-developed \cite{Shin_etal_2017, Seff_etal_2017, Nguyen_etal_2018}. The majority of recent work in this direction has concentrated on implicit generative models \cite{Shin_etal_2017, Seff_etal_2017}, or non-parametric approaches \cite{Milan_etal_2016}. Since these approaches do not possess an inference mechanism, they are unlikely to be useful for subsequent task or policy learning. Furthermore, none of the existing approaches explicitly investigate meaningful sharing of latent representations between environments.


\section{Framework}
\label{sec:framework}

\subsection{Problem formalisation}
\label{sec:formalisation}
We assume that there is an a priori unknown set $\mathcal{S} = \{s_1, s_2, ..., s_K\}$ of $K$ environments which, between them, share a set $\mathcal{Z} = \{z_1, z_2, ..., z_N\}$ of $N$ independent data generative factors. We assume $\vz \sim \mathcal{N}(\mathbf{0},~ \mathbf{I})$. Since we aim to model piece-wise stationary data, it is reasonable to assume $s \sim \operatorname{Cat}(\pi_{1,\ldots,K})$, where $\pi_k$ is the probability of observing environment $s_k$. Two environments may use the same generative factors but render them differently, or they may use a different subset of factors altogether. Given an environment $s$, and an environment-dependent subset $\mathcal{Z}^s \subseteq \mathcal{Z}$ of the ground truth generative factors, it is possible to synthesise a dataset of images $\vx^s \sim p(\cdot|\vz^s, s)$. In order to keep track of which subset of the $N$ data generative factors is used by each environment $s$ to generate images $\vx^s$, we introduce an environment-dependent mask $\va^s$ with dimensionality $|\va| = N$, where $a^s_n=1$ if $z_n \in \mathcal{Z}^s$ and zero otherwise. Hence, we assume $\va^s \sim \operatorname{Bern}(\omega_{1,\ldots,N}^s)$, where $\omega_n^s$ is the probability that factor $z_n$ is used in environment $s$. This leads to the following generative process (where ``$\odot$'' is element-wise multiplication):
\begin{equation}
\begin{gathered}
    \vz \sim \mathcal{N}(\mathbf{0},~ \mathbf{I}), \qquad
        s \sim  \operatorname{Cat}(\pi_{1,\ldots,K}), \qquad    
        \va^s \sim \operatorname{Bern}(\omega_{1,\ldots,N}^s), \\
    \vz^s = \va^s \odot \vz, \qquad 
        \vx^s \sim p(\cdot ~|~ \vz^s, s)
\label{eq:representation-form}
\end{gathered}
\end{equation}

Intuitively, we assume that the piece-wise stationary observed data $\vx$ can be split into \emph{clusters} (environments $s$) (note evidence for similar experience clustering from the animal literature \cite{auchter2017reconsolidation}). Each cluster has a set of \emph{standard coordinate axes} (a subset of the generative factors $\vz$ chosen by the latent mask $\va^s$) that can be used to parametrise the data in that cluster (\cref{fig:network}A). Given a sequence $\vx = (\vx^{s_1},\ \vx^{s_2},\ \ldots)$ of datasets generated according to the process in \cref{eq:representation-form}, where $s_k \sim p(\vs)$ is the $k$-th sample of the environment, the aim of life-long representation learning can be seen as estimating the full set of generative factors $\mathcal{Z}\ \approx \ \bigcup_k~ q(\vz^{s_k}|\vx^{s_k})$ from the environment-specific subsets of $\vz$ inferred on each stationary data cluster $\vx^{s_k}$. Henceforth, we will drop the subscript $k$ for simplicity of notation.



\subsection{Inferring the data generative factors}
\label{sec:mdl}
Observations $\vx^s$ cannot contain information about the generative factors $z_n$ that are not relevant for the environment $s$. Hence, we use the following form for representing the data generative factors:
\begin{equation}
\label{eq:factor-environment-separation}
q(\vz^s|\vx^s) = \va^s ~ \odot ~ \mathcal{N}(\mu(\vx),~ \sigma(\vx)) + (1-\va^s) ~ \odot ~ \mathcal{N}(\mathbf{0},~ \mathbf{I}).
\end{equation}
Note that $\mu$ and $\sigma$ in \cref{eq:factor-environment-separation} depend only on the data $\vx$ and not on the environment $s$. This is important to ensure that the semantic meaning of each latent dimension $z_n$ remains consistent for different environments $s$. We model the representation $q(\vz^s|\vx^s)$ of the data generative factors as a product of independent normal distributions to match the assumed prior $p(\vz) \sim \mathcal{N}(\mathbf{0},~ \mathbf{I})$. 

In order to encourage the representation $q(\vz^s|\vx^s)$ to be semantically meaningful, we encourage it to capture the generative factors of variation within the data $\vx^s$ by following the MDL principle. We aim to find a representation $\vz^s$ that minimises the reconstruction error of the input data $\vx^s$ conditioned on $\vz^s$ under a constraint on the quantity of information in $\vz^s$. This leads to the following loss function:
\begin{equation}
\label{eq:mdl-loss}
\L_{\text{MDL}}(\phi, \theta) = \utext{\E_{\vz^s \sim q_\phi(\cdot|\vx^s) } [-\log~ p_\theta(\vx~|~\vz^s, s)]}{Reconstruction error}\ +\  \gamma\,|\utext{\KLpq{q_\phi(\vz^s|\vx^s)}{p(\vz)}}{Representation capacity} -  \utext{C\vphantom{q_\theta}}{Target}|^2
\end{equation}
The loss in \cref{eq:mdl-loss} is closely related to the \betavae \cite{Higgins_etal_2017} objective $\L = \E_{\vz \sim q_\phi(\cdot|\vx) } [-\log~ p_\theta(\vx|\vz)] + \beta~ \KLpq{q_\phi(\vz|\vx)}{p(\vz)}$, which uses a Lagrangian to limit the latent bottleneck capacity, rather than an explicit target $C$. It was shown that optimising the \betavae objective helps with learning a more semantically meaningful disentangled representation $q(\vz|\vx)$ of the data generative factors \cite{Higgins_etal_2017}. However, \cite{Burgess_etal_2017} showed that progressively increasing the target capacity $C$ in \cref{eq:mdl-loss} throughout training further improves the disentanglement results reported in \cite{Higgins_etal_2017}, while simultaneously producing sharper reconstructions. Progressive increase of the representational capacity also seems intuitively better suited to continual learning where new information is introduced in a sequential manner. Hence, \ourawesomemodel optimises the objective function in \cref{eq:mdl-loss} over a sequence of datasets $\vx^s$. This, however, requires a way to infer $s$ and $\va^s$, as discussed next.

\subsection{Inferring the latent mask}
\label{sec:latent-mask}
Given a dataset $\vx^s$, we want to infer which latent dimensions $z_n$ were used in its generative process (see \cref{eq:representation-form}). This serves multiple purposes: 1) helps identify the environment $s$ (see next section); 2) helps ignore latent factors $z_n$ that encode useful information in some environment but are not used in the current environment $s$, in order to prevent retraining and subsequent catastrophic forgetting; and 3) promotes latent sharing between environments. Remember that \cref{eq:mdl-loss} indirectly optimises for $\E_{\vx^s} [ q_\phi(\vz^s|\vx^s) ] \approx p(\vz)$ after training on a dataset $s$. If a new dataset uses the same generative factors as $\vx^s$, then the marginal behaviour of the corresponding latent dimensions $z_n$ will not change. On the other hand, if a latent dimension encodes a data generative factor that is irrelevant to the new dataset, then it will start behaving atypically and stray away from the prior. We capture this intuition by defining the \emph{atypicality} score $\alpha_n$ for each latent dimension $z_n$ on a batch of data $\vx^s_{\text{batch}}$:
\begin{equation}
\label{eq:atypicality_kl}
\alpha_n = \KLpq{ \textstyle \E_{\vx^s_{\text{batch}}} [ \ q_\phi(z_n^s|\vx^s_{\text{batch}})\ ] \ }{\ p(z_n)}.
\end{equation}
The atypical components are unlikely to be relevant to the current environment, so we mask them out:
\begin{equation}
a_n^s = 
\begin{cases}
 & 1 \text{, if } \alpha_n < \lambda \\ 
 & 0 \text{, otherwise }
\end{cases}
\label{eq:5}
\end{equation}
where $\lambda$ is a threshold hyperparameter  (see \cref{sec:environment-index,sec:hyper_appendix} for more details). Note that the uninformative latent dimensions $z_n$ that have not yet learnt to represent any data generative factors, i.e. $q_\phi(z_n|\vx^s_n) = p(z_n)$, are automatically unmasked in this setup. This allows them to be available as spare latent capacity to learn new generative factors when exposed to a new dataset. Fig.~\ref{fig:traversals} shows the sharp changes in $\alpha_n$ at dataset boundaries during training.

\subsection{Inferring the environment}
\label{sec:environment}

Given the generative process introduced in \cref{eq:representation-form}, it may be tempting to treat the environment $s$ as a discrete latent variable and learn it through amortised variational inference. However, we found that in the continual learning scenario this is not a viable strategy. Parametric learning is slow, yet we have to infer each new data cluster $s$ extremely fast to avoid catastrophic forgetting. Hence, we opt for a fast non-parametric meta-algorithm motivated by the following intuition. Having already experienced $r$ datasets during life-long learning, there are two choices when it comes to inferring the current one $s$: it is either a new dataset $s_{r+1}$, or it is one of the $r$ datasets encountered in the past. Intuitively, one way to check for the former is to see whether the current data $\vx^s$ seems likely under any of the previously seen environments. This condition on its own is not sufficient though. It is possible that environment $s$ uses a subset of the generative factors used by another environment $\mathcal{Z}^s \subseteq  \mathcal{Z}^t$, in which case environment $t$ will explain the data $\vx^s$ well, yet it will be an incorrect inference. Hence, we have to ensure that the subset of the relevant generative factors $\vz^s$ inferred for the current data $\vx^s$ according to \cref{sec:latent-mask} matches that of the candidate past dataset $t$. Given a batch $\vx^s_{\text{batch}}$, we infer the environment $s$ according to:

 
\begin{equation}
s = 
\begin{cases}
 & \hat{s} \text{\ \ \ \ \ \ \ , \ \ if } \ \E_{\vz^{\hat{s}}} [\ p_\theta(\vx^s_{\text{batch}}|\vz^{\hat{s}}, \hat{s}) \ ]\ \leq \kappa L_{\hat{s}} \ \	\land   \ \ \va^s = \va^{\hat{s}} \\ 
 & s_{r+1} \text{, \ \ otherwise }
\end{cases}
\label{eq:6}
\end{equation}

where $\hat{s} =\arg\max_s \ q(s|\vx^s_{\text{batch}})$ is the output of an auxiliary classifier trained to infer the most likely previously experienced environment $\hat{s}$ given the current batch $\vx^s_{\text{batch}}$, $L_{\hat{s}}$ is the average reconstruction error observed for the environment $\hat{s}$ when it was last experienced, and $\kappa$ is a threshold hyperparameter (see \cref{sec:environment-index} for details).  

\subsection{Preventing catastrophic forgetting}
\label{sec:remembering}
So far we have discussed how \ourawesomemodel integrates knowledge from the current environment into its representation $q_\phi(\vz|\vx)$, but we haven't yet discussed how we ensure that past knowledge is not forgotten in the process. Most standard approaches to preventing catastrophic forgetting discussed in \cref{sec:related_work} are either not applicable to a variational context, or do not scale well due to memory requirements. However, thanks to learning a generative model of the observed environments, we can prevent catastrophic forgetting by periodically \emph{hallucinating} (i.e. generating samples) from past environments using a snapshot of \ourawesomemodel, and making sure that the current version of \ourawesomemodel is still able to model these samples. A similar ``dreaming'' feedback loop was used in \cite{ramapuram2017lifelong, Shin_etal_2017, Seff_etal_2017,Ans_Rousset_1997}.

More formally, we follow the generative process in \cref{eq:representation-form} to create a batch of samples $\vx_{\text{old}} \sim q_{\theta_{\text{old}}}(\cdot| \vz, s_{\text{old}})$ using a snapshot of \ourawesomemodel with parameters $(\phi_{\text{old}},\ \theta_{\text{old}})$ (see \cref{fig:network}C). We then update the current version of \ourawesomemodel according to the following (replacing $_{\text{old}}$ with $'$ for brevity):
\begin{equation}
\label{eq:past-loss}
\L_{\text{past}}(\phi, \theta) = \E_{\vz, s', \vx'} \Big[\ 
    \utext{D[q_\phi(\vz|\vx'),\ q_{\phi'}(\vz'|\vx')]}{Encoder proximity}
    + \utext{D[q_\theta(\vx|\vz,s'),\ q_{\theta'}(\vx'|\vz,s')]}{Decoder proximity}\ \Big],
\end{equation}
where $D$ is a distance between two distributions (we use the Wasserstein distance for the encoder and KL divergence for the decoder). The snapshot parameters get synced to the current trainable parameters $\phi_{\text{old}} \leftarrow \phi$,  $\theta_{\text{old}} \leftarrow \theta$ every $\tau$ training steps, where $\tau$ is a hyperparameter. The expectation over simulators $s_{\text{old}}$ and latents $\vz$ in \cref{eq:past-loss} is done using Monte Carlo sampling (see \cref{sec:environment-index} for details).

\subsection{Model summary}
To summarise, we train our model using a meta-algorithm with both parametric and non-parametric components. The latter is needed to quickly associate new experiences to an appropriate cluster, so that learning can happen inside the current experience cluster, without disrupting unrelated clusters. We initialise the latent representation $\vz$ to have at least as many dimensions as the total number of the data generative factors $|\vz| \geq |\mathcal{Z}| = N$, and the softmax layer of the auxiliary environment classifier to be at least as large as the number of datasets $|\mathcal{S}| = K$. As we observe the sequence of training data, we detect changes in the environment and dynamically update the internal estimate of $r \leq K$ datasets experienced so far according to \cref{eq:6}. We then train \ourawesomemodel by minimising the following objective function:
\begin{equation}
\label{eq:full-loss}
\begin{aligned}
\L(\phi, \theta) & = \utext{ \E_{\vz^s \sim q_\phi(\cdot|\vx^s) )} [-\log p_\theta(\vx|\vz^s, s)]\ +\  \gamma\,|\KLpq{q_\phi(\vz^s|\vx^s)}{p(\vz)} -  C|^2}{MDL on current data} + \\
& \ \ \ \ + \utext{ \E_{\vz,s', \vx'} \Big[\ D[q_\phi(\vz|\vx'),\ q_{\phi'}(\vz'|\vx')] + D[q_\theta(\vx|\vz,s'),\ q_{\theta'}(\vx'|\vz,s')]\ \Big]. }{``Dreaming'' feedback on past data}
\end{aligned}
\end{equation}
\section{Experiments}
\label{sec:experiments}
\begin{figure}
    \centering
    \includegraphics[width=1.0\textwidth]{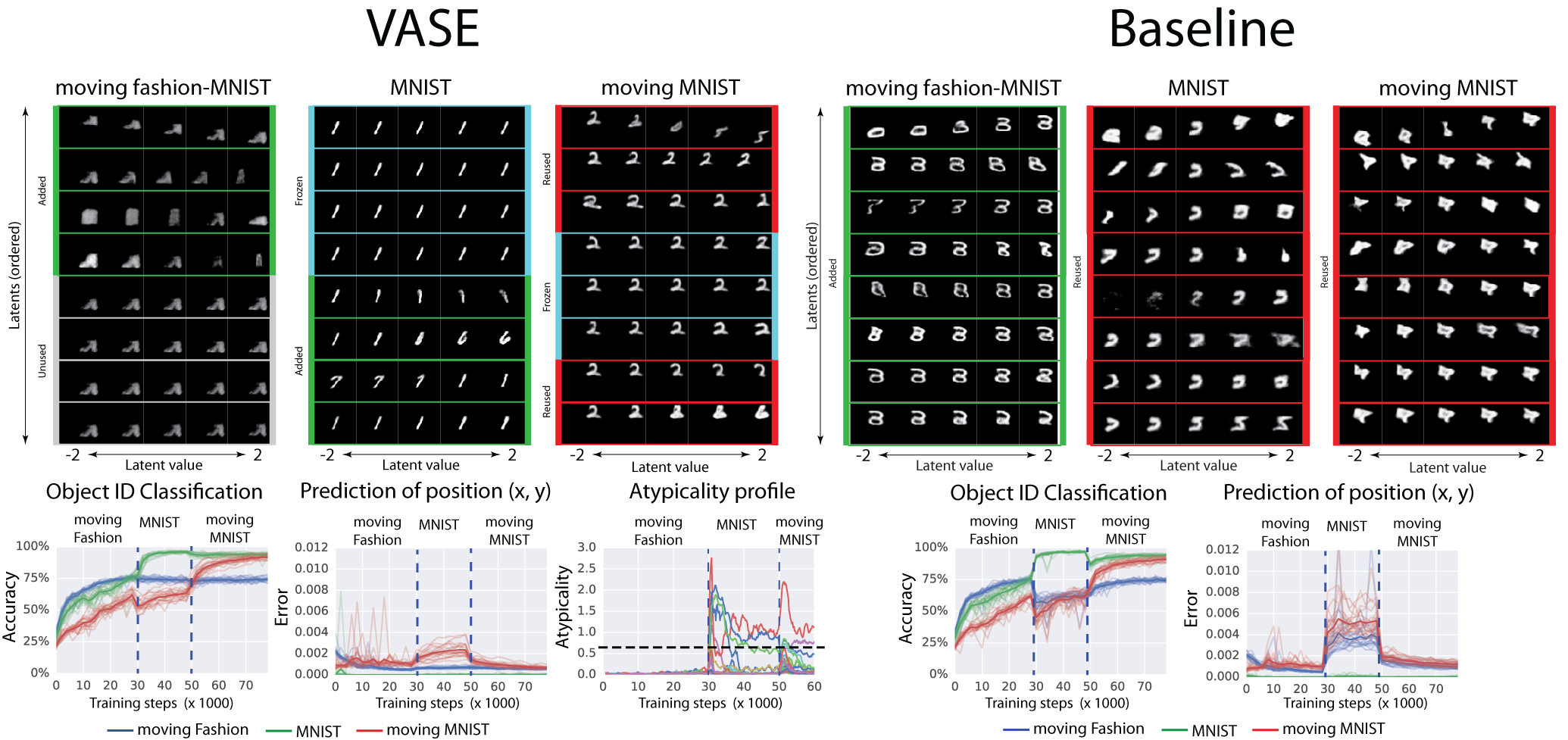}
    \caption{
        We compare \ourawesomemodel to a CCI-VAE baseline. Both are trained on a sequence of three datasets: moving fashion MNIST (\flying \fmnist) $\to$ MNIST $\to$ \flying MNIST. \textbf{Top}: latent traversals at the end of training seeded with samples from the three datasets. The value of each latent $z_n$ is traversed between -2 and 2 one at a time, and the corresponding reconstructions are shown. Rows correspond to latent dimensions $z_n$, columns correspond to the traversal values. Latent use progression throughout training is demonstrated in colour. \textbf{Bottom}: performance of MNIST and \fmnist object classifiers and a position regressor trained on the latent space $\vz$ throughout training. Note the relative stability of the curves for \ourawesomemodel compared to the baseline. The atypicality profile shows the values of $\alpha_n$ through training (different colours indicate different latent dimensions), with the threshold $\lambda$ indicated by the dashed black line.
    }
    \label{fig:traversals}
\end{figure}

\paragraph{Continual learning with disentangled shared latents}
First, we qualitatively assess whether \ourawesomemodel is able to learn good representations in a continual learning setup. We use a sequence of three datasets: (1) a moving version of Fashion-MNIST \cite{Xiao_etal_2017} (shortened to \flying \fmnist), (2) MNIST \cite{LeCun_etal_1998}, and (3) a moving version of MNIST (\flying MNIST). During training we expect \ourawesomemodel to detect shifts in the data distribution and dynamically create new experience clusters $s$, learn a disentangled representation of each environment without forgetting past environments, and share disentangled factors between environments in a semantically meaningful way. Fig.~\ref{fig:traversals} (top) compares the performance of \ourawesomemodel to that of Controlled Capacity Increase-VAE (CCI-VAE) \cite{Burgess_etal_2017}, a model for disentangled representation learning with the same architecture as \ourawesomemodel but without the modifications introduced in this paper to allow for continual learning.
It can be seen that unlike \ourawesomemodel, CCI-VAE forgot \flying \fmnist at the end of the training sequence. Both models were able to disentangle position from object identity, however, only \ourawesomemodel was able to meaningfully share latents between the different datasets - the two positional latents are active for two moving datasets but not for the static MNIST. \ourawesomemodel also has \flying \fmnist- and MNIST-specific latents, while CCI-VAE shares all latents between all datasets. \ourawesomemodel use only 8/24 latent dimensions at the end of training. The rest remained as spare capacity for learning on future datasets. 

\begin{figure}
    \centering
    \includegraphics[width=1.0\textwidth]{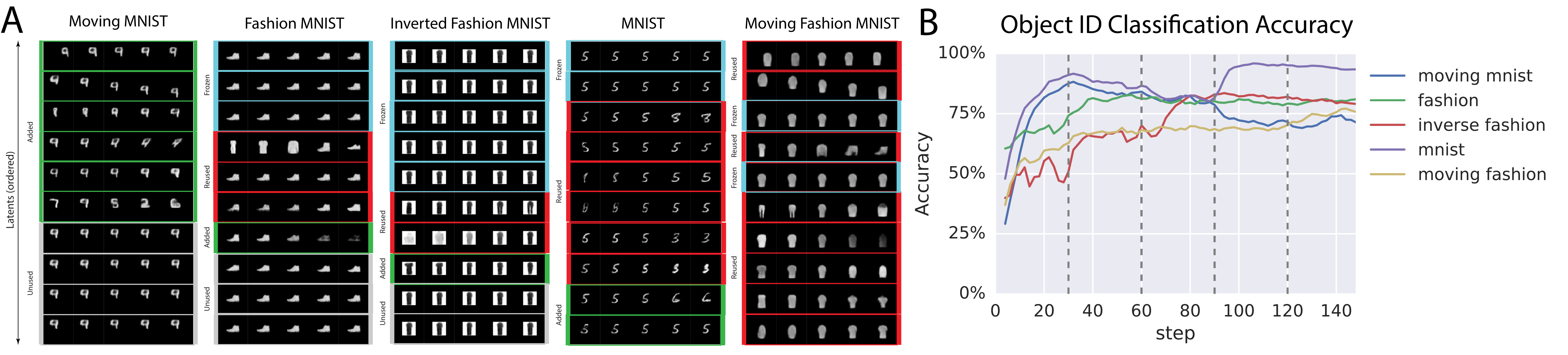}
    \caption{
         Latent traversals (\textbf{A}) and classification accuracy (\textbf{B}) (both as in \cref{fig:traversals}) for \ourawesomemodel trained on a sequence of \flying MNIST $\to$ \fmnist $\to$ inverse \fmnist $\to$ MNIST $\to$ \flying \fmnist. See \cref{fig:5datasets_traverse} for larger traversals.
    }
    \label{fig:5datasets}
\end{figure}

\paragraph{Learning representations for tasks} 
We train object identity classifiers (one each for \flying \fmnist and MNIST) and an object position regressor on top of the latent representation $\vz \sim q_\phi(\vz| \vx)$ at regular intervals throughout the continual learning sequence. Good accuracy on these measures would indicate that at the point of measurement, the latent representation $\vz$ contained dataset relevant information, and hence could be useful, e.g. for subsequent policy learning in RL agents. Figure~\ref{fig:traversals} (bottom) shows that both \ourawesomemodel and CCI-VAE learn progressively more informative latent representations when exposed to each dataset $s$, as evidenced by the increasing classification accuracy and decreasing mean squared error (MSE) measures within each stage of training. However, with CCI-VAE, the accuracy and MSE measures degrade sharply once a domain shift occurs. This is not the case for \ourawesomemodel, which retains a relatively stable representation.

\paragraph{Ablation study} Here we perform a full ablation study to test the importance of the proposed components for unsupervised life-long representation learning: 1) regularisation towards disentangled representations (\cref{sec:mdl}), 2) latent masking (\cref{sec:latent-mask} - \textbf{A}), 3) environment clustering (\cref{sec:environment} - \textbf{S}), and 4) ``dreaming'' feedback loop (\cref{sec:remembering} - \textbf{D}). We use the constraint capacity loss in \cref{eq:mdl-loss} for the disentangled experiments, and the standard VAE loss \cite{Kingma_Welling_2014, Rezende_etal_2014} for the entangled experiments \cite{Higgins_etal_2017}. For each condition we report the average change in the classification metrics reported above, and the average maximum values achieved  (see \cref{sec:quant_forgetting} for details). Table~\ref{tbl:ablation} shows that the unablated \ourawesomemodel (SDA) has the best performance. Note that the entangled baselines perform worse than the disentangled equivalents, and that the capacity constraint of the CCI-VAE framework does not significantly affect the maximal classification accuracy compared to the VAE. It is also worth noting that \ourawesomemodel outperforms the entangled SD condition, which is similar to the only other baseline VAE-base approach to continual learning that we are aware of \cite{ramapuram2017lifelong}. We have also trained \ourawesomemodel on longer sequences of datasets (\flying MNIST $\to$ \fmnist $\to$ inverse \fmnist $\to$ MNIST $\to$ \flying \fmnist) and found similar levels of performance (see \cref{fig:5datasets}).

\begin{table}[t!]
    \begin{center}
    \begin{small}
    \resizebox{\linewidth}{!}{%
        \begin{tabular}{lcc|cc||cc|cc}
        \hline
        & \multicolumn{4}{c||}{\textsc{\textbf{Disentangled}}} & \multicolumn{4}{c}{\textsc{\textbf{Entangled}}} \\
        & \multicolumn{2}{c|}{\textsc{\textbf{Object ID Accuracy}}} & \multicolumn{2}{c||}{\textsc{\textbf{Position MSE}}} & \multicolumn{2}{c|}{\textsc{\textbf{Object ID Accuracy}}} & \multicolumn{2}{c}{\textsc{\textbf{Position MSE}}} \\
        \textsc{Ablation}       &  \textsc{Max (\%)} & \textsc{Change (\%)}  & \textsc{Min (*1e-4)} & \textsc{Change (*1e-4)} &  \textsc{Max (\%)} & \textsc{Change (\%)}  & \textsc{Min (*1e-4)} & \textsc{Change (*1e-4)}  \\
        \hline
        \textsc{-}                & 88.6 ($\pm$0.4) & -15.2 ($\pm$2.8)  &   3.5 ($\pm$0.05) & 24.8 ($\pm$13.5) & 91.8 ($\pm$0.4) & -12.1 ($\pm$0.8) & 4.2 ($\pm$0.7) & 10.5 ($\pm$2.6)  \\
        \textsc{S}                & 88.9 ($\pm$0.5) & -13.9 ($\pm$1.9)  &   3.4 ($\pm$0.05) & 22.5 ($\pm$12.2) & 91.7 ($\pm$0.4) & -12.2 ($\pm$0.03) & 4.5 ($\pm$0.8) & 10.9 ($\pm$3.1)  \\
        \textsc{D}               & 88.6 ($\pm$0.3)  & -14.4 ($\pm$1.9)  &   3.3 ($\pm$0.04) & 21.4 ($\pm$4.9) & 91.8 ($\pm$0.4) & -12.4 ($\pm$0.7) & 4.3 ($\pm$0.7) & 11.7 ($\pm$3.2)  \\        
        \textsc{A}                & 86.7 ($\pm$1.9) & -24.5 ($\pm$1.0)  &   3.3 ($\pm$0.04) & 67.6 ($\pm$107.0) & 88.6 ($\pm$0.3) & -19.7 ($\pm$0.5) & 4.5 ($\pm$0.7) & 47.1 ($\pm$26.2)   \\
        \textsc{SA}              & 87.1 ($\pm$1.8)  & -28.1 ($\pm$0.08)  &  3.3 ($\pm$0.04) & 78.9 ($\pm$109.0) & 89.9 ($\pm$1.3) & -18.3 ($\pm$0.4) & 4.8 ($\pm$0.7) & 41.8 ($\pm$20.6)  \\       
        \textsc{DA}              & 86.3 ($\pm$2.5)  & -25.2 ($\pm$0.5)  &   3.3 ($\pm$0.04) & 72.2 ($\pm$90.0) & 88.8 ($\pm$0.3) & -19.4 ($\pm$0.4) & 4.6 ($\pm$0.7) & 40.2 ($\pm$19.2)  \\
        \textsc{SD}             & 88.3 ($\pm$0.3)  & -12.9 ($\pm$1.9)  &   3.4 ($\pm$0.05) & 20.0 ($\pm$3.5) & 91.4 ($\pm$0.3) & -11.7 ($\pm$0.6) & 4.3 ($\pm$0.5) & 11.6 ($\pm$1.9)  \\
        \hline
        \textsc{\textbf{\ourawesomemodel} (SDA)}             & 88.6 ($\pm$0.4)   & \textbf{-5.4 ($\pm$0.3)} &    3.2 ($\pm$0.03) & \textbf{3.0 ($\pm$0.2)} & 91.5 ($\pm$0.1) & -6.5 ($\pm$0.7) & 4.2 ($\pm$0.4) & 3.9 ($\pm$1.1)  \\
        \hline
        \hline
        \end{tabular}
    }
    \caption{Average change in classification accuracy/MSE and maximum/minimum average accuracy/MSE when training an object/position classifier/regressor on top of the learnt representation on the \flying \fmnist $\to$ MNIST $\to$ \flying MNIST sequence. We do a full ablation study of \ourawesomemodel, where D - dreaming feedback loop, S - cluster inference $q(s|\vx^s)$, and A - atypicality based latent mask $\va^s$ inference. We compare two versions of our model - one that is encouraged to learn a disentangled representation through the capacity increase regularisation in \cref{eq:mdl-loss}, and an entangled VAE baseline ($\beta=1$). The unablated disentangled version of \ourawesomemodel (SDA) has the best performance.}
    \label{tbl:ablation}
    \end{small}
    \end{center}
\end{table}

\begin{figure}[t]
    \centering
    \includegraphics[width=1.0\textwidth]{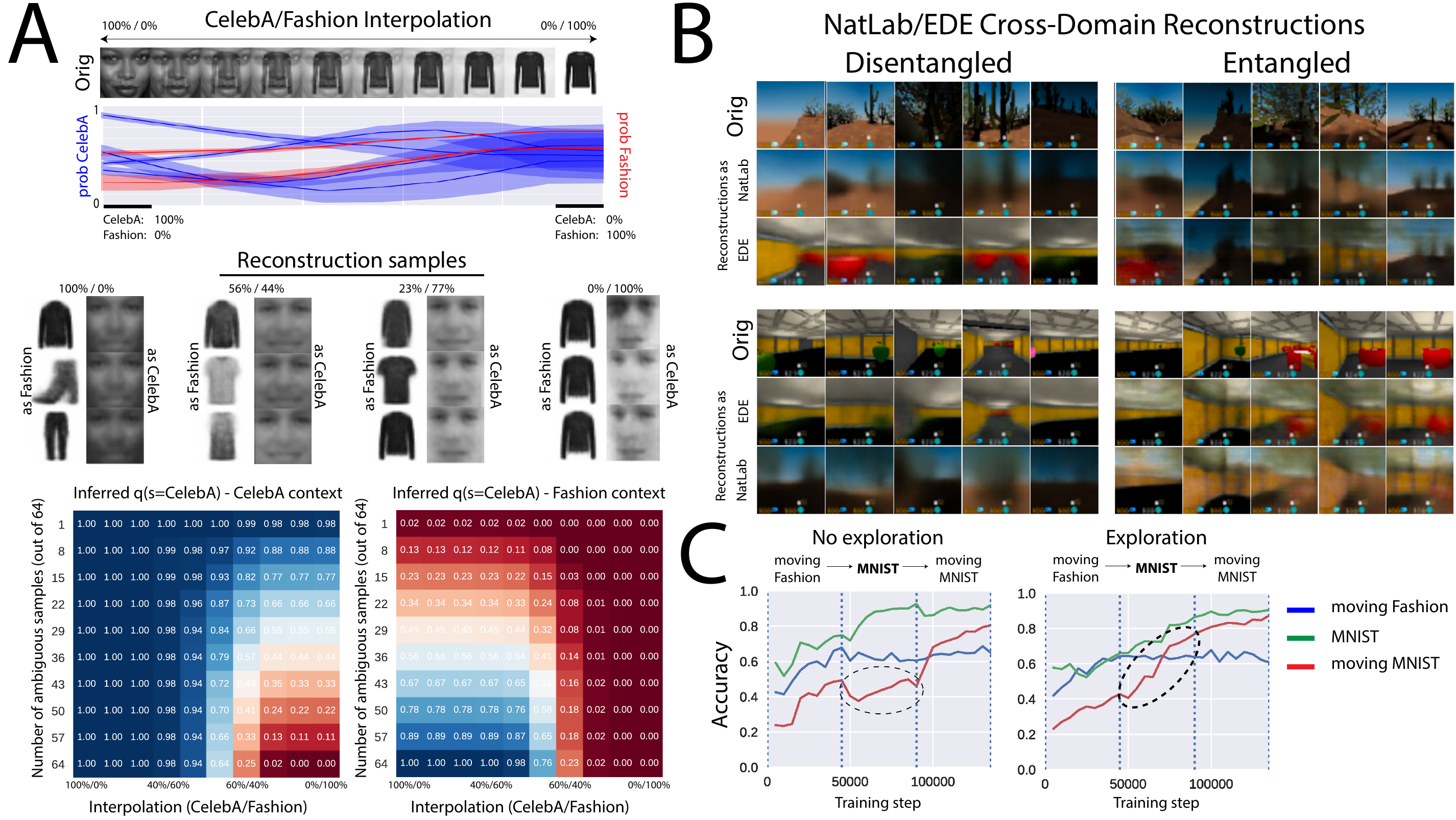}
    \caption{
    \textbf{A} Top: Ambiguous input examples created by using different interpolation weights between samples from CelebA and \fmnist, and corresponding inferred parameters $\mu$ (y axis) and $\sigma$ (light colour range) of $q_\phi(\vz|\vx)$; red corresponds to \fmnist-specific latents, blue to CelebA-specific latents. Middle: Reconstruction samples $p_\theta(\vx^s|\vz^s, s)$ for different levels of ambiguity conditioned on either dataset. Bottom: Inferred $q_\psi(s=\text{CelebA}$ given different levels of input ambiguity (x axis) and different number of ambiguous vs real data samples (y axis) for the two datasets. \ourawesomemodel deals well with ambiguity, shows context-dependent categorical perception and uncertainty within its inferred representation parameters. \textbf{B} Cross-domain reconstructions on NatLab (outdoors) or EDE (indoors) DM Lab levels. The disentangled \ourawesomemodel finds semantic homologies between the two datasets (e.g. cacti $\to$ red objects). The entangled \ourawesomemodel only maps lower level statistics. \textbf{C} Imagination-based exploration allows \ourawesomemodel to imagine the possibility of \flying MNIST digits during static MNIST training by using position latents acquired on \flying \fmnist. This helps it learn a \flying MNIST classifier during static MNIST training without ever seeing real translations of MNIST digits.
    }
    \label{fig:experiments}
\end{figure}

\paragraph{Dealing with ambiguity}
Natural stimuli are often ambiguous and may be interpreted differently based on contextual clues. Examples of such processes are common, e.g. visual illusions like the Necker cube \cite{Necker_1832}, and may be driven by the functional organisation and the heavy top-down influences within the ventral visual stream of the brain \cite{Balasz_etal_1993, Przybyszewski_1998}. To evaluate the ability of \ourawesomemodel to deal with ambiguous inputs based on the context, we train it on a CelebA \cite{Liu_etal_2015} $\to$ inverse \fmnist sequence, and test it using ambiguous linear interpolations between samples from the two datasets (\cref{fig:experiments}A, first row). To measure the effects of ambiguity, we varied the interpolation weights between the two datasets. To measure the effects of context, we presented the ambiguous samples in a batch with real samples from one of the training datasets, varying the relative proportions of the two. Figure~\ref{fig:experiments}A (bottom) shows the inferred probability of interpreting the ambiguous samples as CelebA $q_\phi(s=\text{celebA}|\vx)$. \ourawesomemodel shows a sharp boundary between interpreting input samples as \fmnist or CelebA despite smooth changes in input ambiguity. Such \emph{categorical perception} is also characteristic of biological intelligence \cite{Etcoff_Magee_1992, Freedman_etal_2001, Liu_Jagadeesh_2008}. The decision boundary for categorical perception is affected by the context in which the ambiguous samples are presented. \ourawesomemodel also represents its uncertainty about the ambiguous inputs by increasing the inferred variance of the relevant latent dimensions (\cref{fig:experiments}A, second row).

\paragraph{Semantic transfer}
Here we test whether \ourawesomemodel can learn more sophisticated cross-domain latent homologies than the positional latents on the \flying MNIST and \fmnist datasets described above. Hence, we trained \ourawesomemodel on a sequence of two visually challenging DMLab-30 \footnote{\url{https://github.com/deepmind/lab/tree/master/game\_scripts/levels/contributed/dmlab30\#dmlab-30}} \cite{deepmind_lab} datasets: the Exploit Deferred Effects (EDE) environment and a randomized version of the Natural Labyrinth (NatLab) environment (Varying Map Randomized). While being visually very distinct (one being indoors and the other outdoors), the two datasets share many data generative factors that have to do with the 3D geometry of the world (e.g. horizon, walls/terrain, objects/cacti) and the agent's movements (first person optic flow). Hence, the two domains share many semantically related factors $\vz$, but these are rendered into very different visuals $\vx$. We compared cross-domain reconstructions of \ourawesomemodel and an equivalent entangled VAE ($\beta=1$) baseline. The reconstructions were produced by first inferring a latent representation based on a batch from one domain, e.g. $\vz^{\text{NatLab}} = q_\phi(\cdot|\vx^{\text{NatLab}})$, and then reconstructing them conditioned on the other domain $\vx^{\text{xRec}} = q_\theta(\cdot|\vz^{\text{NatLab}}, s^{\text{EDE}})$. Fig.~\ref{fig:experiments} shows that \ourawesomemodel discovered the latent homologies between the two domains, while the entangled baseline failed to do so. \ourawesomemodel learnt the semantic equivalence between the cacti in NatLab and the red objects in EDE, the brown fog corresponding to the edge of the NatLab world and the walls in EDE (top leftmost reconstruction), and the horizon lines in both domains. The entangled baseline, on the other hand, seemed to rely on the surface-level pixel statistics and hence struggled to produce meaningful cross-domain reconstructions, attempting to match the texture rather than the semantics of the other domain. See \cref{sec:additional_results} for additional cross-domain reconstructions, including on the sequence of five datasets mentioned earlier.

\paragraph{Imagination-driven exploration}
Once we learn the concept of moving objects in one environment, it is reasonable to imagine that a novel object encountered in a different environment can also be moved. Given the ability to act, we may try to move the object to realise our hypothesis. We can use such imagination-driven exploration to augment our experiences in an environment and let us learn a richer representation. Notice however, that such imagination requires a compositional representation that allows for novel yet sensible recombinations of previously learnt semantic factors. We now investigate whether \ourawesomemodel can use such imagination-driven exploration to learn better representations using a sequence of three datasets: \flying \fmnist $\to$ MNIST $\to$ \flying MNIST. During the first \flying \fmnist stage, \ourawesomemodel learns the concepts of position and \fmnist sprites. It also learns how to move the sprites to reach the imagined states $z^*$ by training an auxiliary policy (see \cref{sec:self-motivation} for details). It can then use this policy to do an imagination-based augmentation of the input data on MNIST by imagining MNIST digits in different positions and transforming the static sprites correspondingly using the learnt policy. Hence, \ourawesomemodel can imagine the existence of \flying MNIST before actually experiencing it. Indeed, \cref{fig:experiments}C shows that when we train a \flying MNIST classifier during the static MNIST training stage, the classifier is able to achieve good accuracy in the \emph{imagination-driven exploration} condition, highlighting the benefits of imagination-driven data augmentation.

\section{Conclusions}

We have introduced \ourawesomemodel, a novel approach to life-long unsupervised representation learning that builds on recent work on disentangled factor learning \cite{Higgins_etal_2017, Burgess_etal_2017} by introducing several new key components. Unlike other approaches to continual learning, our algorithm does not require us to maintain a replay buffer of past datasets, or to change the loss function after each dataset switch. In fact, it does not require any a priori knowledge of the dataset presentation sequence, since these changes in data distribution are automatically inferred. We have demonstrated that \ourawesomemodel can learn a disentangled representation of a sequence of datasets. It does so without experiencing catastrophic forgetting and by dynamically allocating spare capacity to represent new information. It resolves ambiguity in a manner that is analogous to the categorical perception characteristic of biological intelligence. Most importantly, \ourawesomemodel allows for semantically meaningful sharing of latents between different datasets, which enables it to perform cross-domain inference and imagination-driven exploration. Taken together, these properties make \ourawesomemodel a promising algorithm for learning representations that are conducive to subsequent robust and data-efficient RL policy learning.

\newpage
\small
\bibliography{bibliography}
\bibliographystyle{abbrv} 

\newpage
\appendix
\section{Supplemental Details}

\subsection{Model details}
\label{sec:model-details}

\paragraph{Encoder and decoder} For the encoder we use a simple convolutional network with the following structure: 
\texttt{
conv 64 $\to$ conv 64 $\to$ conv 128$\to$ conv 128$\to$ fc 256}, where \texttt{conv n\_filters} is a $4\times4$ convolution with \texttt{n\_filters} output filters, ReLU activations and stride 2, and similarly \texttt{fc n\_out} is a fully connected layer with \texttt{n\_out} units. The output of the fully connected layer is given to a linear layer that outputs the mean $\mu_\enc(\vx)$ and log-variance $\log \sigma^2_\enc(\vx)$ of the encoder posterior $q_\phi(\vz|\vx) \sim N(\mu_\enc(\vx), \sigma^2_\enc(\vx))$. The decoder network receives a sample $\vz \sim q_\phi(\vz|\vx)$ from the encoder and outputs the parameters of a distribution $p_\theta(\vx|\vz,s)$ over $\vx$. We use the transpose of the encoder network, but we also feed it the environment index $s$ by first encoding it with a one-hot encoding (of size \texttt{max\_environments}, which is a hyperparameter), and then concatenating it to $\vz$.  For most of the experiments, we use a product of independent Bernoulli distributions (parametrised by the mean) for the decoder distribution $p_\theta(\vx|\vz, s)$. In the DM Lab experiments we use instead a product of Gaussian distributions with fixed variance. We train the model using Adam \cite{Kingma_Ba_2014} with a fixed learning rate 6e-4 and batch size 64.

\paragraph{Environment inference network} We attach an additional fully connected layer to the last layer of the encoder (gradients to the encoder are stopped). Given an input image $\vx$, the layer outputs a softmax distribution $q_\psi(s|\vx)$ over \texttt{max\_environments} indices, which tries to infer the most likely index $s$ of the environment from which the data is coming, assuming the environment was already seen in the past. Notice that we always know the (putative) index of the current data ($\hat{s}$ in \Cref{eq:6}, also see \Cref{sec:environment-index}), so that we can always train this layer to associate the current data to the current index. However, to avoid catastrophic forgetting we also need to train on hallucinated data from past environments. Assuming $\hat{s}$ is the current environment and $m$ is the total number of environments seen until now, the resulting loss function is given by:
\[
\L_{\text{env}} = \utext{\E_{\vx}[-\log(q_\psi(\hat{s}|\vx))]}{Classification loss on current data} + \utext{\E_{\hat{s} \neq s<m}\E_{\vx' \sim p_{\theta'}(\vx'|\vz', s)}[-\log q_{\psi}(s|\vx')]}{Classification loss on hallucinated data},
\]
where the hallucinated data $\vx'$ in the second part of the equation is generated according to \Cref{sec:remembering}, and the expectation over $s$ is similarly done through Monte Carlo sampling.

\subsection{Extra algorithm implementation details}
\label{sec:environment-index}

\paragraph{Atypical latent components} The atypicality $\alpha_n$ of the component $z_n$ on a batch of samples $\vx_1,\ldots,\vx_B$ is computed using a KL divergence from the marginal posterior over the batch to the prior according to \Cref{eq:atypicality_kl}. In practice it is not convenient to compute this KL divergence directly. Rather, we observe that the marginal distribution of the latent samples $\frac{1}{B} \sum_{b=1}^B q_\phi(z_n|\vx_b)$ is approximately Gaussian. We exploit this by fitting a Gaussian to the latent samples $\vz$ and then computing in closed-form the KL-divergence between this approximation of the marginal and the unit Gaussian prior $p(\vz) = \mathcal{N}(0, 1)$. 

Recall from \Cref{sec:latent-mask} that we deem a latent component $z_n$ to be active ($a_n=1$) whenever it is typical, that is, if $\alpha_n < \lambda$. However, since the atypicality is computed on a relatively small batch of $B$ samples, $\alpha_i$ may be a noisy estimate of atypicality. Hence we introduce the following filtering: we set $\alpha_n=1$ if $\alpha_n>\lambda_1$ and $\alpha_n=0$ if $\alpha_n<\lambda_0$, with $\lambda_0 < \lambda_1$. If $\lambda_0 < \alpha_n < \lambda_1$, we leave $\alpha_n$ unchanged.

\paragraph{Used latent components} We say that a factor $z_n$ is not used by the environment $s$ if the reconstruction $p_\theta(\vx|\vz,s)$ does not depend on $z_n$. To measure this, we find the maximum amount of noise we can add to $z_n$ without changing the reconstruction performance of the network. That is, we optimise
\[
\Sigma = \underset{{\Sigma = \diag (\sigma_1,\ldots,\sigma_N)}}{\operatorname{argmin}} \E_{\epsilon \sim \mathcal{N}(0,\sigma)}[-\log p_\theta(\vx|\vz^{\epsilon},s)] - \log |\Sigma|
\]
where $\vz_n^{\epsilon} = (1 - \delta_{nm}) z_n + \delta_{nm} (z_n+\epsilon)$. If $\sigma'_n > T$ for some threshold $T$, we say that $z_n$ is unused. We generally observe that components are either completely unused $\sigma_n = 0$, or else $\sigma_n$ is very large. Therefore, picking a threshold is very easy and the precise value does not matter. We only compare the atypicality masks in \cref{eq:6} for the used latents.

\paragraph{Environment index} Expanding on the explanation in \Cref{sec:environment}, let $L_s(\vx)=\E_{\vz\sim q_\phi(\vz|\vx)}[-\log p_\theta(\vx|\vz^s,s)]$ be the reconstruction loss on a batch $\vx$ of data, assuming it comes from the environment $s$. Let $\tilde{L}_s$ be the average reconstruction loss observed in the past for data from environment $s$. Let $m$ be the number of datasets observed until now. Let $\vu^s$ be a binary vector of used units computed with the method described before.

We run the auxiliary environment inference network (\Cref{sec:model-details}) on each sample from the batch $\vx$ and take the average of all results in order to obtain a probability distribution $q(s|\vx)$ over the possible environment $s$ of the batch $\vx$, assuming it has already been seen in the past. Let $\hat{s}=\arg\max_s q(s|\vx)$ be the most likely environment, which is our candidate for the new environment. If the reconstruction loss $L_{\hat{s}}(\vx)$ (assuming $s=\hat{s}$) is significantly larger (see \Cref{alg:environment}) than the average loss for the environment $\hat{s}$, we decide that the data is unlikely to come from this environment, and hence we allocate a new one. If the reconstruction is good, but some of the used components (given by $\vu$) are atypical, we still allocate a new environment. Otherwise, we assume that the data indeed comes from $\hat{s}$.

\begin{algorithm}
\caption{Infer the environment index $s$ from a batch of data}
\begin{algorithmic}
\State $\hat{s} \gets \arg\max_{s} \E_{\vz\sim q_\phi(\vz|\vx)}[-\log p_\theta(\vx|\vz^s,s)]$
\If {$L_{\hat{s}} > \kappa \tilde{L}_{\hat{s}}$}
    \State $s \gets m + 1$
\ElsIf{$a^{\hat{s}} \odot \vu^s \neq \va(\vx) \odot \vu^{\hat{s}}$}
    \State $s \gets m + 1$
\Else
    \State $s \gets \hat{s}$
\EndIf
\end{algorithmic}
\label{alg:environment}
\end{algorithm}

\subsection{Hyperparameter sensitivity}
\label{sec:hyper_appendix}

\Cref{tbl:hyper} lists the values of the hyperparameters used in the different experiments reported in this paper.

\begin{table}[t!]
    \begin{center}
    \begin{tiny}
    \resizebox{\linewidth}{!}{%
        \begin{tabular}{lcccccc||ccccc}
        \hline
        & \multicolumn{6}{c||}{\textsc{\textbf{Disentangled}}} & \multicolumn{3}{c}{\textsc{\textbf{Entangled}}} \\
        \textsc{Experiment}       &  \textsc{$\gamma$} & \textsc{$C_\text{max}$} & $\delta C$  & \textsc{$\lambda$} & \textsc{$\kappa$} & \textsc{$\tau$} & \textsc{$\lambda$} & \textsc{$\kappa$} & \textsc{$\tau$}\\
        \hline
        \textsc{Ablation Study (150k)} & 100.0 & 35.0 & 6.3e-6  & 0.6 & 1.5 & 500 & & \\
        \textsc{Five Datasets (150k)} & 100.0 & 35.0 & 6.3e-6 & 0.6 & 1.5 & 5000 & & \\
        \textsc{CelebA $\to$ Inverted Fashion (30k)}  & 200.0 & 20.0 & 1.7e-5 & 0.8 & 1.1 &  500 &  & \\
        \textsc{NatLab $\to$ EDE (60k)}  & 200.0 & 25.0 & 1e-5 & 2.0 & 1.1 & 5000 & 20.0 & 1.1 & 5000 \\
        \textsc{Imagination-Driven Exploration (45k)}  & 200.0 & 35.0 & 0.7e-5 & 0.7 & 1.5 & 500 & & & \\
        \hline
        \hline
        \end{tabular}
    }
    \caption{Hyperparameter values used for the experiments reported in this paper. Values in brackets after the experiment name indicate the number of training steps used per dataset.}
    \label{tbl:hyper}
    \end{tiny}
    \end{center}
\end{table}

For all experiments we use \texttt{max\_environments} = 7, and we increase $C$ in \cref{eq:mdl-loss} linearly by $\delta C \cdot C_\text{max}$ per step (starting from 0) until it reaches $C_\text{max}$, at which point we keep $C$ fixed at that value. In the loss function \cref{eq:full-loss}, the dreaming loss was re-weighted, with the full loss being:

\begin{equation}
\label{eq:full-loss-weighted}
\begin{aligned}
\L(\phi, \theta) & = \utext{ \E_{\vz^s \sim q_\phi(\cdot|\vx^s) )} [-\log p_\theta(\vx|\vz^s, s)]\ +\  \gamma\,|\KLpq{q_\phi(\vz^s|\vx^s)}{p(\vz)} -  C|^2}{MDL on current data} + \\
& \ \ \ \ + \utext{ \E_{\vz,s', \vx'} \Big[\ \alpha D[q_\phi(\vz|\vx'),\ q_{\phi'}(\vz'|\vx')] + \beta D[q_\theta(\vx|\vz,s'),\ q_{\theta'}(\vx'|\vz,s')]\ \Big]. }{``Dreaming'' feedback on past data}
\end{aligned}
\end{equation}

The values $\alpha=1000$ and $\beta=20$ were used for all experiments, except for in the hyperparameter sweep.

For the ablation study we ran a hyperparameter search using the full model, and used the best hyperparameters found for all experiments. We list the search ranges and our observed sensitivity to these hyperparameters next: 

\begin{itemize}
  \item $\gamma$ = coefficient for the capacity constraint -- $\{50, 100, 200\}$  -- found not to be very sensitive.
  \item $C_\text{max}$ = final value of $C$  -- $\{20, 35, 50\}$  -- classification accuracy increased significantly for capacity from 20 to 35.
  \item $\lambda$ = atypicality threshold -- $\{0.4, 0.6, 1, 2\}$  -- lower threshold led to more latent freezing. Classification performance was not very sensitive to this.
  \item $\tau$ = update frequency reference network -- $\{500, 1000, 2000, 5000\}$  -- found not to be very sensitive.
  \item $\alpha$ = weight for encoder loss in "dreaming" loop -- $\{10, 20, 40\}$ -- found not to be very sensitive.
  \item $\beta$ = weight for decoder loss in "dreaming" loop -- $\{500, 1000, 2000\}$ -- found not to be very sensitive.
\end{itemize}

\subsubsection{Imagination-based exploration experiments}
\label{sec:self-motivation}

We model a very simple interaction with the environment where the agent can translate the observed object (in our case the input image). The agent is trained as follows: a random $\vz^*$ is sampled from the prior $p(\vz)$. Given an observation $\vx$ from the environment, the agent needs to pick an action $g(\vz^*,\vx)$ (in our case a translation) in such a way that the encoding $\vz \sim q_\phi(\vz|g \cdot \vx)$ of the new image $g \cdot \vx$ is as close as possible to $\vz^*$. That is, we minimise the loss
\[
\L_{\text{agent}} = \E_{\vx \sim p(\vx)}\E_{\vz^* \sim p(\vz)} \E_{\vz \sim q(\vz|g(\vz^*,\vx) \cdot \vx)} \|\vz^* - \vz\|^2.
\]
The agent can then be used to explore the current environment. Given an image $\vx$ from the current environment, we can imagine a configuration $\vz^*$ of the latent factors, and let the agent act on the environment in order to realise the configuration. In this way, we obtain a new image $\vx^* = g(\vz,\vx) \cdot \vx$. We can then add the image $\vx^*$ to the training data, which allows the encoder to learn from a possibly more diverse set of inputs than the inputs $\vx \sim p(\vx)$ observed passively.

The policy network first processes the input image $\vx$ (of size $64\times 64$) though four convolutional $4\times 4$ layers with 16 filters, stride 2 and ReLU activations. The resulting vector is concatenated with the target $\vz^*$, and feed to a 1-hidden layer fully connected network that outputs the parameters of the 2D translation $g(\vx,\vz^*)$ to apply to the image. We use a \texttt{tanh} output to ensure the translation is always in a sensible range. Once these parameters are obtained, the transformation is applied to the image $x$ using a Spatial Transformer Network (STN) \cite{jaderberg2015spatial}, obtaining a translated image $\vx^* = g(\vz^*,\vx) \cdot \vx$. We can now finally compute  the resulting representation $\vz \sim q(\vz|g(\vz^*,\vx) \cdot \vx)$. Notice that the whole operation is fully differentiable, thanks to the properties of the STN. The policy is can now be trained by minimising $\L_{\text{agent}}$ in order to make $\vz$ and $\vz^*$ as close as possible. In our experiments we train the policy while training the main model.

\subsection{Dataset processing}
\paragraph{DM Lab}
We used an IMPALA agent trained on all DM-30 tasks \cite{Espeholt_etal_2018} to generate data. We take observations of this optimal agent (collecting rewards according to the task descriptions explained in \url{https://github.com/deepmind/lab/tree/master/game_scripts/levels/contributed/dmlab30}), on randomly generated episodes of Exploit Deferred Effects and NatLab Varying Map Randomized; storing them as $111 \times 84 \times 3$ RGB tensors.
We crop the right-most 27 pixels out to obtain a $84 \times 84$ image (this retains the most useful portion of the original view), which are finally scaled down to $64 \times 64$ (using \texttt{tf.image.resize\_area}).

\paragraph{CelebA $\to$ Inverse \fmnist}
To make CelebA compatible with \fmnist, we convert the CelebA images to grayscale and extract a patch of size $32\times 32$  centered on the face. We also invert the colours of \fmnist so that the images are black on a white background, and slightly reducing the contrast, in order to make the two datasets more similar, and hence easier to confuse after mixing.

\subsection{Quantifying catastrophic forgetting}
\label{sec:quant_forgetting}
We train on top of the representation $\vz \sim q_\phi(\vz|\vx)$ a simple 2-hidden layers fully connected classifier with 256 hidden units per layer and ReLU activations. At each step while training the representation, we also train a separate classifier on the representation for each environment, using Adam with learning rate 6e-4 and batch size 64. This classifier training step does not update the weights in the main network.

For each ablation type we reported the average classification accuracy (or regression MSE) score obtained by 20 replicas of the model, all with the best set of hyperparameters discovered for the full model. We quantified catastrophic forgetting by reporting the average difference between the maximum accuracy obtained while \ourawesomemodel was training on a particular dataset and the minimum accuracy obtained for the dataset afterwards.

\subsection{Additional results}
\label{sec:additional_results}
\begin{figure}
    \centering
    \includegraphics[width=.48\textwidth]{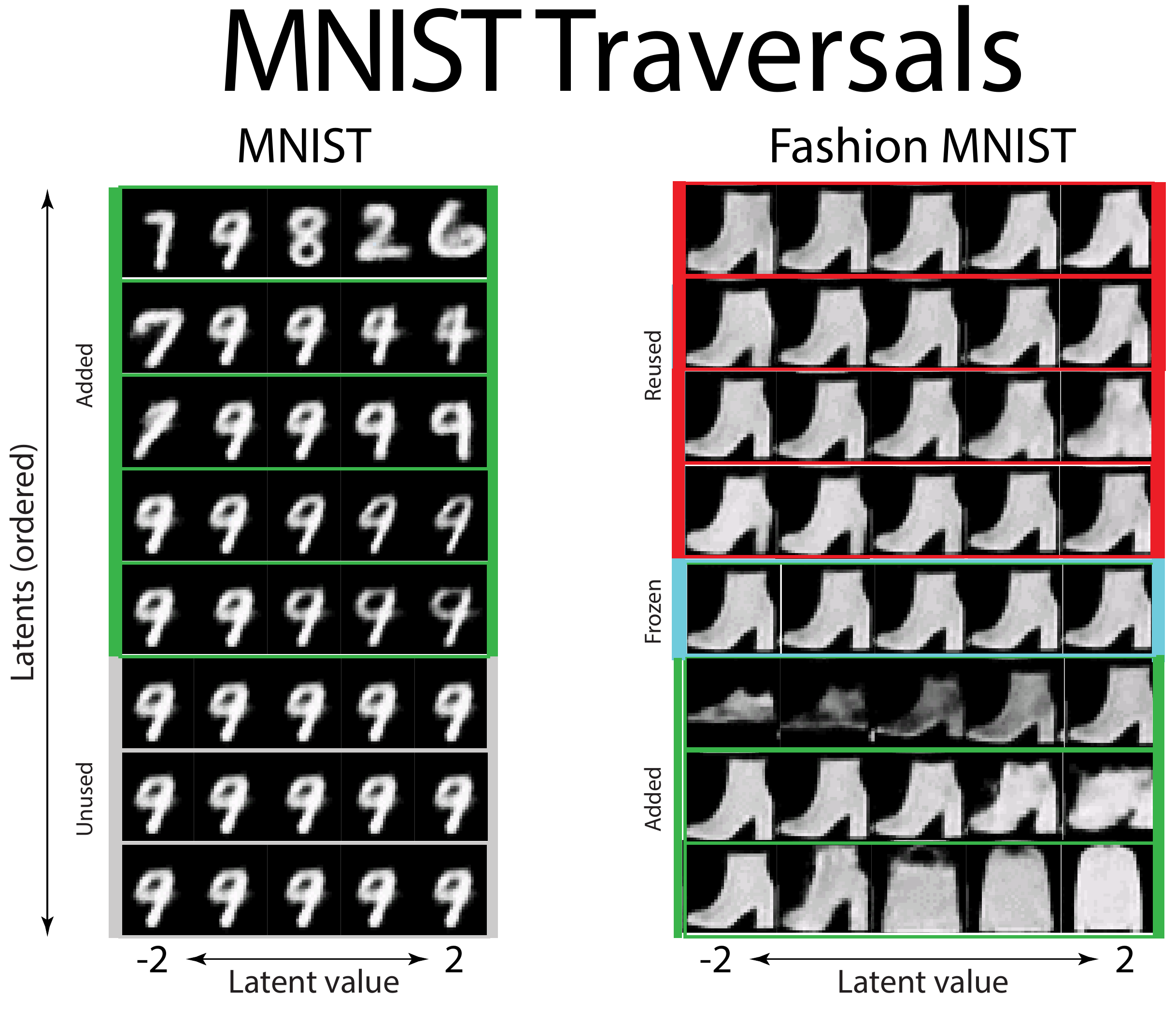}
    \includegraphics[width=.40\textwidth]{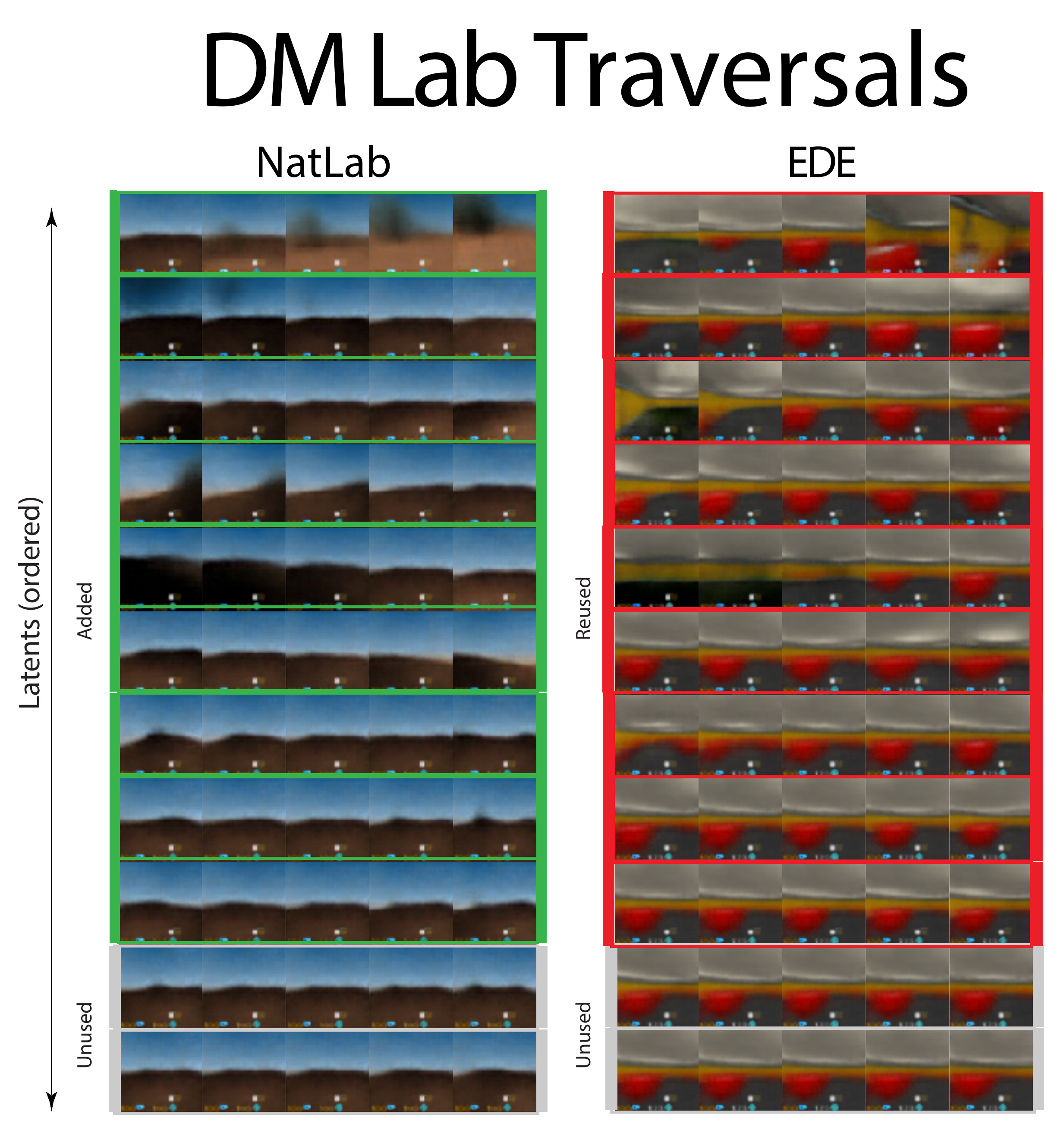}
    \caption{
    Latent traversals for \ourawesomemodel trained on MNIST $\to$ \fmnist $\to$ MNIST, and DM Lab levels NatLab $\to$ EDE.
    }
    \label{fig:mnist_fmnist_traversals}
\end{figure}

\begin{figure}
    \centering
    \includegraphics[width=.95\textwidth]{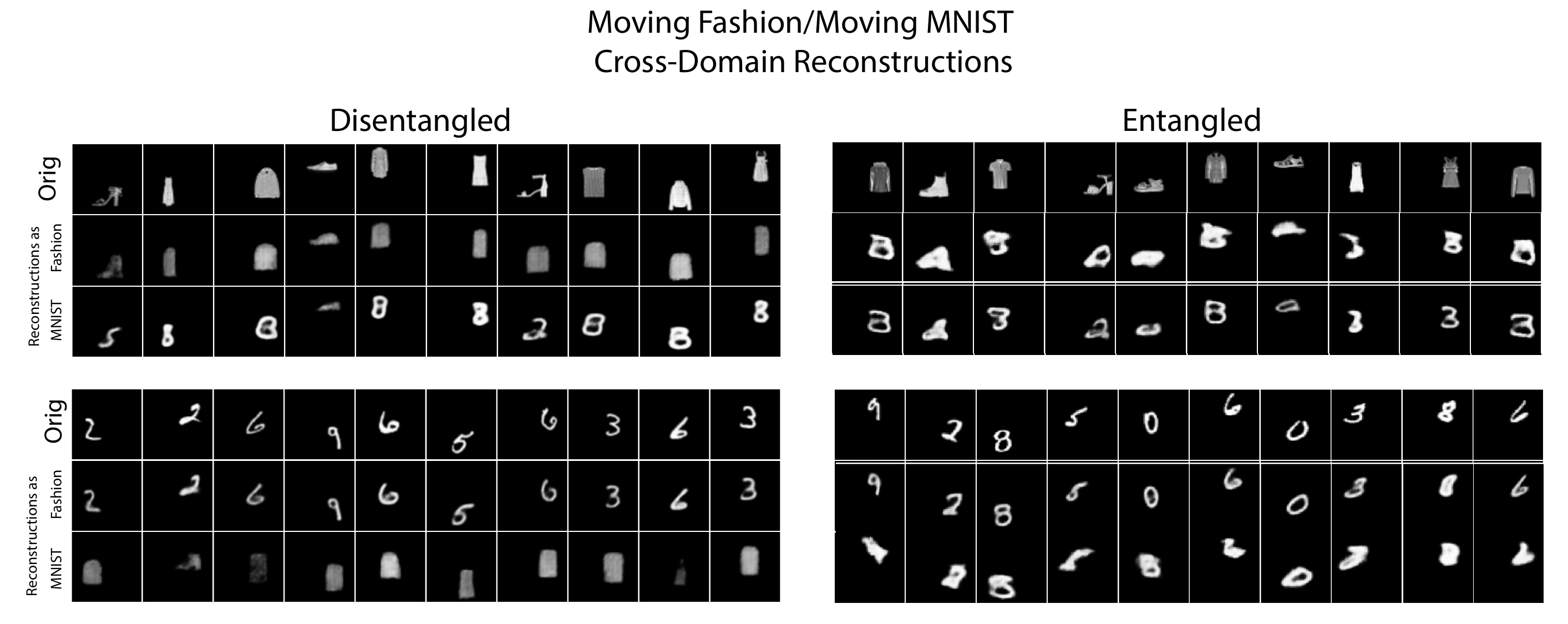}
    \caption{
    Cross-domain reconstructions for the entangled and disentangled versions of \ourawesomemodel (as described in \cref{sec:experiments}) trained on \flying \fmnist $\to$ MNIST $\to$ \flying MNIST. We see that the entangled baseline forgets \flying \fmnist by the end of training. 
    }
    \label{fig:cross_domain_moving_mnist}
\end{figure}

\begin{figure}
    \centering
    \includegraphics[width=1.0\textwidth]{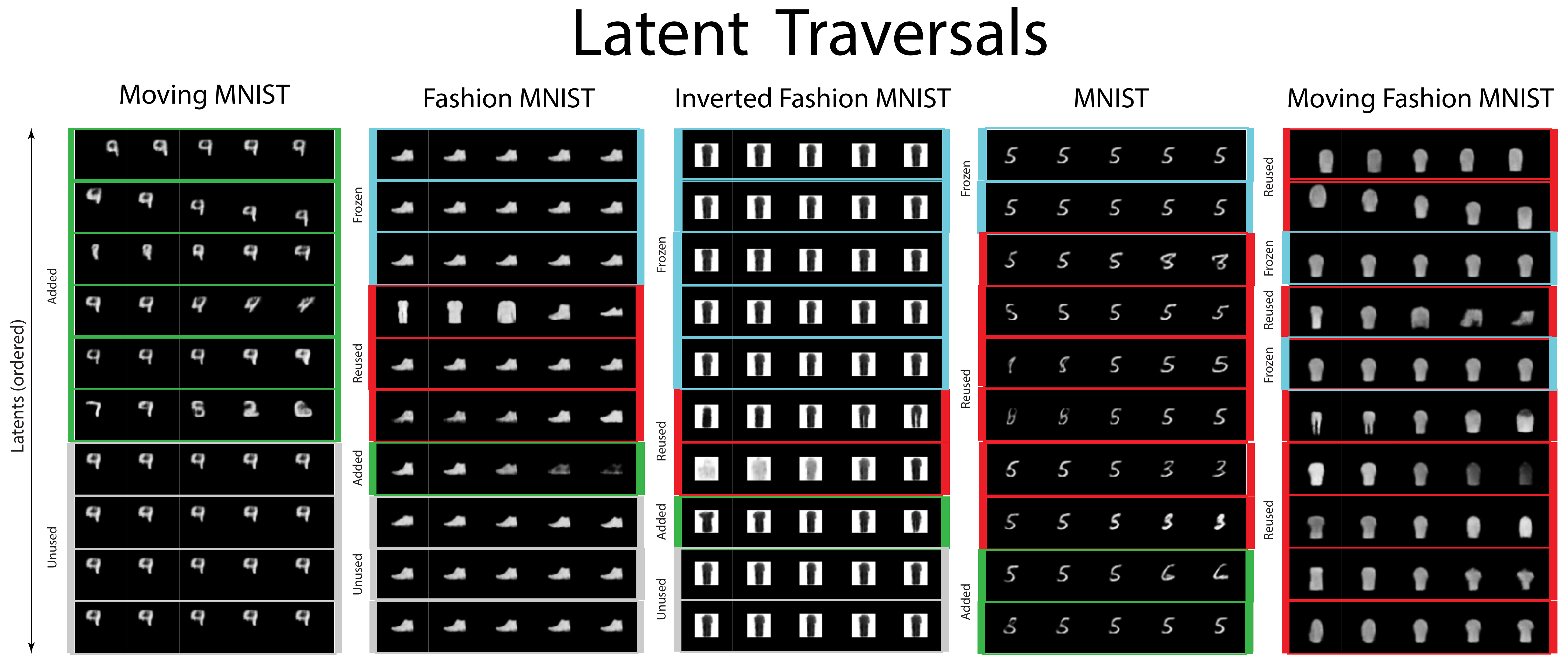}
    \caption{
        Latent traversals for \ourawesomemodel trained on a sequence of \flying MNIST $\to$ \fmnist $\to$ inverse \fmnist $\to$ MNIST $\to$ \flying \fmnist.
    }
    \label{fig:5datasets_traverse}
\end{figure}

\begin{figure}
    \centering
    \includegraphics[width=.95\textwidth]{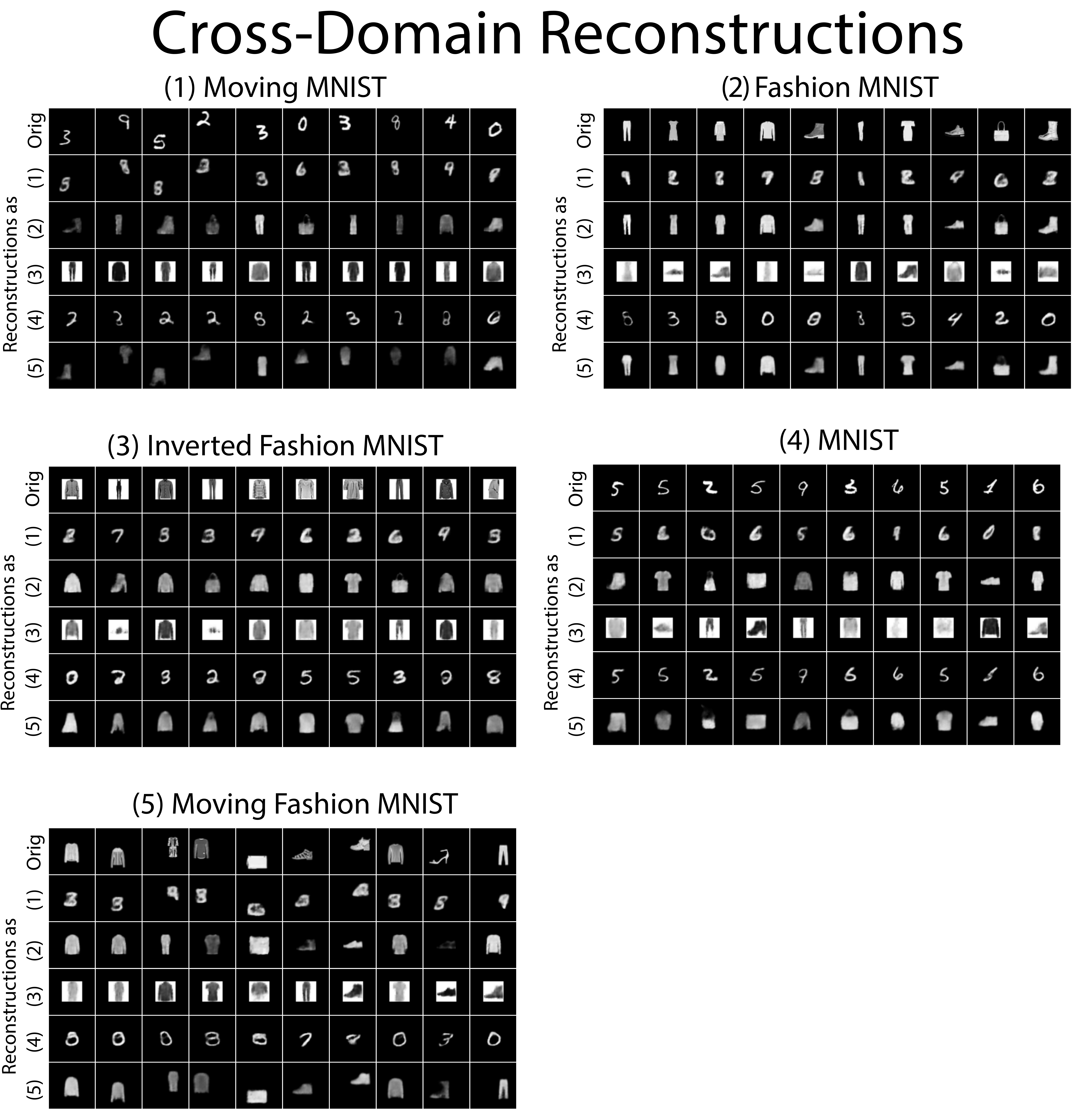}
    \caption{
    Cross-domain reconstructions for the \ourawesomemodel trained on \flying MNIST $\to$ \fmnist $\to$ inverted \fmnist $\to$ MNIST $\to$ \flying \fmnist.
    }
    \label{fig:5dataset_cross_traversals}
\end{figure}

We present additional experimental results and extra plots for the experiments reported in the main paper here. Fig.~\ref{fig:mnist_fmnist_traversals} and \cref{tbl:ablation_appendix} show latent traversals and quantitative evaluation results for an ablation study on \ourawesomemodel trained on the MNIST $\to$ \fmnist $\to$ MNIST sequence. Fig.~\ref{fig:mnist_fmnist_traversals} also shows traversals for \ourawesomemodel trained on the DM Lab levels NatLab $\to$ EDE. This is the model reported in \cref{sec:experiments}. Fig.~\ref{fig:cross_domain_moving_mnist} shows cross-dataset reconstructions for \ourawesomemodel trained on the \flying \fmnist $\to$ MNIST $\to$ \flying MNIST sequence described in the ablation study in \cref{sec:experiments}. Figs.~\ref{fig:5datasets_traverse}-\ref{fig:5dataset_cross_traversals} shows latent traversals and cross-dataset reconstructions for \ourawesomemodel trained on the \flying MNIST $\to$ \fmnist $\to$ inverted \fmnist $\to$ MNIST $\to$ \flying \fmnist sequence described in the main text.

\begin{table}[t!]
    \begin{center}
    \begin{small}
    \resizebox{\linewidth}{!}{%
        \begin{tabular}{lcc||cc}
        \hline
        & \multicolumn{2}{c||}{\textsc{\textbf{Disentangled}}} & \multicolumn{2}{c}{\textsc{\textbf{Entangled}}} \\
        \textsc{Configuration}       &  \textsc{Avg. Decrease (\%)} & \textsc{Avg. Max (\%)}  & \textsc{Avg. Decrease (\%)} & \textsc{Avg. Max (\%)}  \\
        \hline
        \textsc{DA}               & -7.9  & 90.5 & -12.1 & 90.9  \\
        \textsc{SD}               & -2.2  & 91.0 & -4.3 & \textbf{92.1}  \\
        \textsc{S}               & -3.9  & 90.8 & -8.4 &  91.5 \\
        \textsc{A}               &  -9.6 & 90.2 & -10.1 & 91.7  \\
        \textsc{SA}               & -5.9  & 90.0 & -10.3 & 91.0  \\
        \textsc{-}               & -4.4  & 91.1 & -6.6 &  92.6 \\
        \textsc{D}               &  -6.0 & 90.5 & -6.9 & 91.4  \\
        \hline
        \textsc{\textbf{\ourawesomemodel} (SDA)}   & \textbf{-0.9} & 90.3 & -2.5 & 91.2 \\
        \hline
        \hline
        \end{tabular}
    }
    \caption{Average drop in classification accuracy and maximum average accuracy when training an object classifier on top of the learnt representation on the MNIST $\to$ \fmnist $\to$ MNIST sequence. We do a full ablation study of \ourawesomemodel, where D - dreaming feedback loop, S - cluster inference $q(s|\vx^s)$, and A - atypicality based latent mask $\va^s$ inference. We compare two versions of our model - one that is encouraged to learn a disentangled representation through the capacity increase regularisation in \cref{eq:mdl-loss}, and an entangled VAE baseline ($\beta=1$). The unablated disentangled version of \ourawesomemodel (SDA) has the best performance.}
    \label{tbl:ablation_appendix}
    \end{small}
    \end{center}
\end{table}

\end{document}